# AMV : Algorithm Metadata Vocabulary

Biswanath Dutta[1][0000-0003-3059-8202], Jyotima Patel[2]

DRTC, Indian Statistical Institute, Bangalore -560059, India
[1]`bisu@drtc.isibang.ac.in`
[2]`jyotima@drtc.isibang.ac.in`

**Abstract.** Metadata vocabularies are used in various domains of study. It provides an in-depth description of the resources. In this work, we develop Algorithm Metadata Vocabulary (AMV), a vocabulary for capturing and storing the metadata about the algorithms (a procedure or a set of rules that is followed step-by-step to solve a problem, especially by a computer). The snag faced by the researchers in the current time is the failure of getting relevant results when searching for algorithms in any search engine. AMV is represented as a semantic model and produced OWL file, which can be directly used by anyone interested to create and publish algorithm metadata as a knowledge graph, or to provide metadata service through SPARQL endpoint. To design the vocabulary, we propose a well-defined methodology, which considers real issues faced by the algorithm users and the practitioners. The evaluation shows a promising result.

## 1 Introduction

An Algorithm is a bit by bit game plan to tackle any issue. It very well may be composed as flowcharts, pseudo-codes, or in some programming language. It tends to be utilized to take care of an unpredictable issue by separating it into steps [7]. The motivation behind an algorithm is to get the essential embodiment of the issue so we can get it and make fitting moves. Algorithms are broadly utilized in the current advanced time, for example, top hashtags on Twitter, moving recordings on YouTube, and so on are naturally identified and shown to the clients by different complex algorithms [18,19]. Thinking about the significance of algorithms, they can be called plans for imperative computational cycles.

In this work, we propose a metadata vocabulary for catching and putting away the metadata about algorithms. There are uncountable algorithms present in every area (e.g., Computer Science, Mathematics), which makes it hard for analysts, academicians, application designers, and so on to discover, recognize, select, and reuse them. In the current situation, we for the most part search the algorithms utilizing the web indexes like Google, Bing, and Yahoo. As we know, it is elusive and select the necessary Algorithm. Since for the most part the algorithms are published as a feature of the logical writing. Consequently, discovering them on the Web includes least two stages: first recover the important writing and afterward by physically preparing them to track down the pertinent algorithm. Additionally, in the current situation, we typically search them by the subject. This confines the inquiry, makes the pursuit interaction a complex and tedious undertaking. We can't scan the algorithms by their properties for instance by information structure, issue, time intricacy, and so on. Our suggestion is, to make the algorithm search more powerful and customized, they ought to be treated as an autonomous computerized object, like the exploration article, research information, and in the new time the ontologies. Given the meaning of the algorithms as advanced resources, they ought to be very much depicted and managed. There ought to be devoted storehouse for keeping up algorithms like the repositories accessible for the previously mentioned objects, for instance, ontology repositories, like BioPortal (https://bioportal.bioontology.org/), AgroPortal (http://agroportal.lirmm.fr/), data repositories, like https://datadryad.org/, https://nssdc.gsfc.nasa.gov/, etc.In regard to algorithm, we discovered two well appriciated initiatives, like The Stony Brook Algorithm Repository (https://algorist.com/algorist.html) and AlgoWiki (https://wiki.algo.is/). Both the storehouses give data to the different existing algorithms including their source. The significant issues with the two of them are, there is no search facility. In the Stony Brook repository, we can browse the algorithm by issues and by the language. In AlgoWiki, the famous algorithms are enrolled. The more insights regarding these archives are given in segment 2. One of the essential necessities to work with the item search is the metadata. We need metadata vocabulary to empower the object portrayal and organization. Further, we need committed and explicit metadata vocabulary for this reason. The presently accessible metadata vocabularies (e.g., Dublin Core (DC), DCAT, LOM) for different computerized resources are adequately not to depict algorithms. algorithms have a few qualities which are novel and can't be caught by any such normalized vocabularies. They might be utilized to catch the most nonexclusive data, like title, creator, date, identifier, and so on. In our investigation, we were unable to track down any current activities towards making a metadata vocabulary especially for the depiction of the algorithm. A committed metadata vocabulary towards algorithms will likewise help in distinguishing proof and choice of the ideal algorithm. Utilizing a clear cut metadata vocabulary, we can make a storehouse which will empower the consistent correlation among algorithms and makes understanding them simple. Algorithm Metadata Vocabulary (AMV) is the first of its sort. It is an exertion towards making algorithm metadata as a FAIR information [10] (Findable, Accessible, Interoperable, and Reusable). AMV can be utilized by the algorithm storehouse engineers and administrators, and furthermore can be utilized for annotating the writing on algorithms. The primary commitment of the current work is it gives a shiny new metadata metadata vocabulary. Plus, we have addressed AMV as a semantic model and delivered OWL file, which can be straightforwardly utilized by anybody intrigued to make and distribute algorithm metadata as a knowledge graph [27], or to provide metadata service through SPARQL endpoint [26]. It merits referencing here that AMV has not been made in a disconnected manner. All things considered, it has been made by reusing the current vocabularies, like DC, schema.org, and so forth wherever applicable. The genuine worries of the professionals in the domain are also taken into consideration.

The rest of the paper is organized as follows: section 2 discusses the current state of the algorithm repositories; section 3 provides the guiding principles and methodology to build AMV; section 4 deals with the top-level facets, which describes the outlook of an algorithm defined to design the AMV; section 5 deals with the metadata model describing the various classes and properties and also the vocabularies that we (re)used to in AMV; section 6 contains the evaluation of the model with the help of a survey and SPARQL queries; section 7 discusses some relevant and related state-of-the-art works. Finally, section 8 concludes the paper, and also provides future research directions.

## 2 Current State of Algorithm Repositories

As referenced above, there are two algorithm repositories accessible and they are the AlgoWiki and the Stony Brook Algorithm Repository. Here, we momentarily depict them.

The algowiki is an online reference book of algorithms, it contains a rundown of algorithms in a wiki page organized in sequential request [12]. When you select your ideal algorithm and snap on it, the site diverts you to a page where the algorithm is clarified and the pseudo-code is given, the diverted site is a Wikipedia page or any tech site. The Stony Brook Algorithm Repository chiefly centers around statistical surveying to discover which algorithmic issues are utilized for the most part in applications primarily zeroing in on the field of combinatorial algorithms and algorithm engineering [13]. The aftereffect of the exploration done showed that the principle motivation behind the visitors in the archive was to look for executions of algorithms that tackle the problem they are keen on.

The stony brook repository has just some predefined key Algorithms and those algorithms are assembled in 3 classifications: By language, By problem, and By Algorithm types. On the off chance that one ticks on a algorithm the page contains the portrayal of the problem, algorithm, the input description, and the different executions of similar algorithm to take care of different issues is given alongside the ratings[11]. The stony brook repository hasn't been updated since 2008.

The presently existing stores have different disadvantages, for instance searching is not an easy task as they are HTML based and just works with browsing. The minimal annotations act as a bar in keyword searching. Consequently, a metadata metadata vocabulary devoted to algorithms is needed for thorough depiction, appropriate portrayal of algorithms, which will work with better administration and support of the repository. The metadata vocabulary will likewise empower the individual store designers to keep up the algorithm explicit information all the more proficiently. The presence of standard components in the metadata vocabulary will work with better utilization and reuse of algorithms, accordingly making the different algorithm stores interoperable as there will be consistency regarding the components used to explicitly portray the algorithms.

Given the above perception, we have contemplated the different algorithms and extracted the properties showed by them for better order and furthermore to give better depiction of them. We have additionally contemplated the different existing metadata vocabularies to fabricate an establishment for our work as definite in the accompanying area.

## 3 Methodology

Here, we momentarily examine the plan approach of AMV alongside the guiding priciples. Like some other metadata vocabulary advancement action, the AMV likewise began with indicating its motivation and tracking down the connected assets including the quest for related existing vocabularies. The AMV advancement measure comprises of top-down and bottom-up perspectives and each further comprises of various advances. Previously, explaining the AMV improvement steps, we initially give a bunch of guiding principles as follows initially gave in [5, 8]. These core values, particularly the initial three one are pertinent when we mean to reuse the current vocabularies to make another one.

The guiding principles are
1. Precedence to the metadata vocabularies particular towards algorithm rather than more generic ones
2. Conflict among the classes must not be there (e.g., Disjoint class)

3. The considered vocabulary must be a recommendation of standard organizations, such as World Wide Web consortium, Dublin Core, etc.
4. Maintain balance between adequacy and essential while creating a metadata vocabulary.
5. Reusing the existing vocabularies for standardizing element names.
6. Interoperability of the vocabulary should be maintained.
7. The metadata elements used should be distinct, with understandable description.
8. The reuse of the resource must be reinforced by the vocabulary.

### 3.1 AMV Approach

*Top-down Approach (deductive approach).* This methodology requests continuing from a calculated level to a solid level, i.e., the wide perspective on AMV ought to be taken in thought. We work on the theoretical level by referencing the high level aspects for which we draw up the few highlights of the asset to be depicted. In this work, the center asset is a algorithm. Subsequent to depicting the high level aspects we further investigate them and restricted them down to the various classes and sub-classes level as explained in segment 4

*Bottom-up Approach (inductive approach).* This methodology requests continuing from a solid level to a theoretical level. In this methodology, we contemplated and distinguished the properties of a algorithm. We attempted to dissect how the properties extricated are connected with the classes referenced in the hierarchical methodology. To be precise in our work, we continued bit by bit as portrayed in Figure 1. The steps mentioned below give an idea of how we created the vocabulary from scratch.

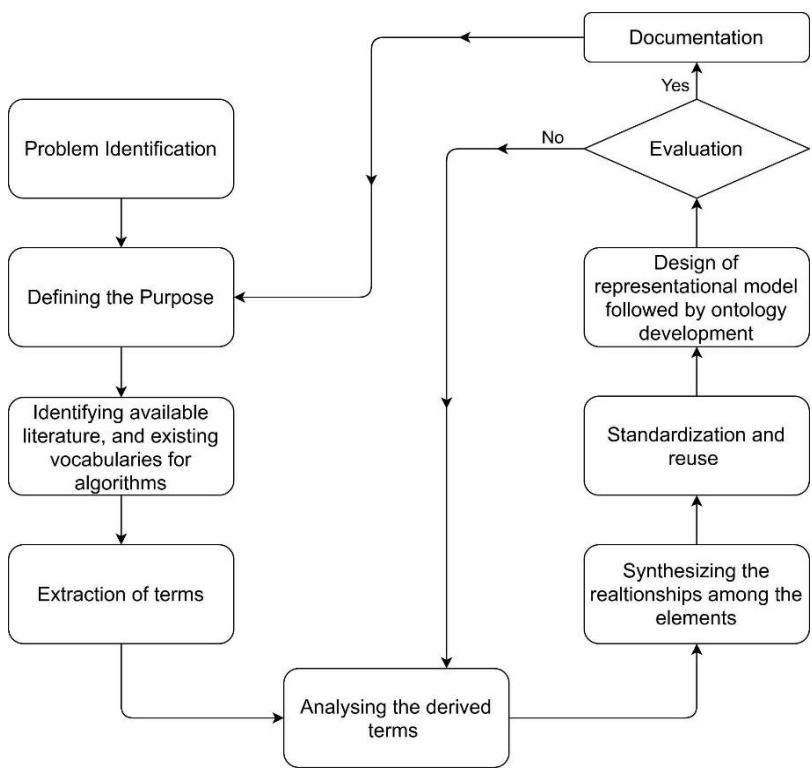

**Fig. 1.** Steps to create AMV

*Step 0: Problem identification*
In this progression we distinguish the requirements confronted while looking for algorithms. This is finished by analyzing the current algorithm archives. This progression depicts how the current storehouses come up short on the appropriate hunt and recovery office as they are HTML based and just work with browsing. The absence of appropriate filtering criteria while searching. One primary limitation confronted while looking for a particular algorithm is that multiple occasions that algorithms dwell in the writing just and are not accessible on the web or any store.

*Step 1: Defining the purpose*
The progression portrays the reason for our work. As talked about beforehand, the fundamental motivation behind the work is to regard algorithms as a digital entity and make a metadata vocabulary explicit to algorithms and normalize our metadata vocabulary by (re)using the current pertinent vocabularies used to portray other computerized elements. The ultimate objective is to make algorithms explicit information as a FAIR information. Explaining the reason characterized above into a bunch of competency questions. A portion of the inquiries are:

1. Which sorting algorithms use data structure array?
2. Which language independent algorithm belongs to subject mathematics and solves combinatorial problems?
3. Retrieve the list of algorithms that are implemented in python or java and have less than 7 steps?
4. List the greedy algorithms for which 'while loop' is the best suited loop type?
5. Retrieve all the sorting algorithms which are expressed in the form of flowchart?

*Step 2: Reviewing the available literature and existing vocabularies*
In this progression we considered the accessible writing managing the algorithms, qualities of algorithms and furthermore the metadata vocabularies that could be in part used to portray algorithms. A portion of the center writing that we counseled are given in Table 1. Moreover, we likewise evaluated a few existing metadata vocabularies for their possible use in AMV. The further insights concerning the utilization of the current vocabularies are given in the succeeding segments.

Table. 1. Lists the crucial literature that we consulted for our work

| Serial no. | Literature consulted |
|---|---|
| 1. | Voevodin, V., Antonov, A., & Dongarra, J. (2016). Why is it Hard to Describe Properties of Algorithms? Procedia Computer Science, 101, 4-7. doi:10.1016/j.procs.2016.11.002 |
| 2. | Cormen, T. H., Leiserson, C. E., Rivest, R. L., & Stein, C. (2009). Introduction to algorithms. Cambridge (Mass.): MIT Press. |
| 3. | Skiena, S. S.(1998). The algorithm design manual. New York: Springer. |

*Step 3: Extraction of terms*
Recognizing the regular properties of a algorithm that apply to most algorithms, similar to the Algorithm Type, creator, and so on. In extricating the properties, we initially centered around the nonexclusive terms which normally all digital resources have, for instance, creator, date, and publisher. We additionally alluded to the two algorithm archives as referenced above, i.e., the Stony Brook Algorithm Repository and AlgoWiki. We have embraced the components from them, particularly from Stony Brook repository including its Algorithm grouping pattern by Problem. The inside and out investigations of the algorithm assisted us with separating properties, similar to time complexity, edge case, data structure, implementation, and so on. In this progression, we zeroed in on the properties explicit to algorithms. For extricating these properties we examined algorithms of different areas, like software engineering, science, and so on

*Step 4: Analysing the derived terms*
.In this step we enlisted, defined, and analysed all the extracted terms. The extracted complex terms were broken into their elemental entities. We analysed them on the basis of their characteristics and then grouped them on the basis of similarity.

*Step 5: Synthesizing the relationships among the elements*
In this step we arrange and synthesize them based on their relationships with other entities. In this step we describe how the entities are related with other entities and discover the concept hierarchy. For example, following the previous step 4, the terms *average case*, best *case*, and *worst case* are grouped together and placed under a more generic term time complexity.

*Step 6: Reusing the existing vocabularies and standardization*
This step is important because integration and (re)use of the existing vocabularies leads to development and reuse of the domain vocabulary. We have integrated concepts from Dublin Core Terms, Schema.org, FOAF(Friend of a Friend), SKOS. DOAP, and DCAT.

*Step 7: Design of representational model followed by Ontology development*
In this step, we first structure and model the domain knowledge produced in the previous step. We model the domain knowledge representing its components, such as classes, properties and their relationships. Followed by development of the formal model using a formal logic language OWL. We used the Protégé [25] ontology editor for designing the ontology. The details provided in section 5.

*Step 8: Evaluation*

This progression includes assessing how the planned metadata metadata vocabulary and the model work progressively. There is no programmed method of assessing the area information. Thus, to assess, we led a survey and the reactions were in support of ourselves. For the model, the reasoners confirmed the design and consistency of the philosophy and furthermore executed a bunch of SPARQL inquiries arranged by changing the competency addresses gave in sync 1. On the off chance that the assessment is effective with every one of the conditions fulfilled, we continue towards the documentation. Assuming the assessment comes up short at a specific advance, we re-try the way toward examining the terms to improve depiction of the algorithm.. For additional subtleties on assessment, see segment 6

*Step 9: Documentation*
Followed by the fruitful assessment, we make a metadata vocabulary particular archive which can be utilized by the client for the consistent utilization of AMV. In future, for refreshing/forming we can continue to stage 1 and follow the strides ahead. The AMV documentation may be accessed from https://isibang.ac.in/ns/amv/

## 4 Top-level Facets

Following the Top-down approach, we found the high level features for the algorithm metadata vocabulary. The different high level aspects give an outline of the algorithm metadata metadata vocabulary showing its different highlights. The high level features were additionally investigated and limited to the different classes of AMV gave in area 5.1. The different high level aspects are determined keeping algorithms as the focal point of the examination. We investigated the algorithm from the different casing of reference and distinguished seven high level aspects as follows:

*General* – describes the general features of an algorithm, for instance, the algorithm type, algorithm notation (the form of expression), etc.

*Authority* –describes the organization and/ or the person responsible for the algorithm.

*Rights* – the licenses under which an algorithm is protected, and the rights held on it.

*Iterations and solution* – describes the various concepts used in the algorithm and the iterations used, for instance, mathematical property, loop types, etc.

*Coverage* – describes the algorithm problem and the discipline it belongs to.

*Environment* – describes the environment an algorithm needs when converted into a program, for instance, programming languages, operating systems, etc.

*Resource* – describes the various type of resources, such time, space, etc. that are required to run an algorithm.

## 5 AMV Metadata Model

The AMV metadata vocabulary has been expressed as a formal semantic model. We have developed and published it as an OWL ontology. Anyone (e.g., the repository managers, developers, users) interested in creating and publishing algorithm metadata as a knowledge graph can directly download the AMV ontology and populate it with the data. Then the knowledge graph can be either published on Linked Data Cloud or even can be used for providing direct access to the data through SPARQL endpoint. The present AMV 1.0 ontology(available at https://isibang.ac.in/ns/amv) consists of 27 classes, 24 object properties and 46 data properties. Each of these are further described in the following sections.

### 5.1 Classes

The AMV consists of 27 classes (a class is a collective name that is used for a set of individuals sharing common properties).As discussed previously, the classes are derived following the identification of various aspects of the algorithm and they are depicted in the Table 1 as top-level facets (from left, the column 1). The derived classes are organized by interlinking them through the hierarchical and associative relations (aka object properties). The hierarchical relation rdfs:subClassOf is used to organize the classes in their class and sub-class relations. Table 1 depicts the classes and sub-classes (column 2) and their corresponding example instances (column 3). The classes are interlinked using the various object properties as discussed in the following section 5.2. Besides, the classes are also defined using a set of data properties as described in section 5.3.

**Table. 2.** Depicts the AMV top-level facets along with the classes and sub-classes and some examples of class instances. The classes and sub-classes are separated by "," and also the sub-classes are expresses in *italics*.

| Top-level facets | Class name | Example of class instance |
|---|---|---|
| General | Algorithm | Binary Search, Merge Sort |
| | AlgorithmType | Dynamic, recursive |
| | FormOfExpression | Flowchart, Pseudo code |

| Authority | Agent, *Organization*, *Person* | Organization related with the algorithm and the person associated with it. |
|---|---|---|
| Right | RightsStatement, *LicenseDocument* | Creative Commons, Copyleft, Apache |
| Iteration and solution | MathematicalProperty | Linear algebra, Theory of sets |
|  | LoopTypes | For, while, do-while |
| Coverage | Discipline | Graph, mathematics |
|  | Problem, *CombinatorialProblem*, *ComputationalGeometryProblem*, *DataStructuresProblem*, *NumericalProblem*, *SetAndStringProblem*, *GraphProblem*, *HardProblem*, *PolynomialTimeProblem* | Convex Hull, Range Search, Independent Set, Linear Programming |
| Environment | ProgrammingLanguage | C, Fortran, Java |
|  | OperatingSystem | Linux, Windows |
|  | Implementation | BSD Sort, N Sort |
|  | DataStructure | Array,Lists,HashTables |
| Resource | AlgorithmAnalysisFunction | Constant, logarithmic |

**5.2 Object Property**
An object property relates an entity with another entity. The entities belonging to the same or different classes are connected using the object properties. AMV consists of 24 object properties which include creator, subject, problem type, source, license, form of expression, loop construct, has implementation, and so forth. The domain and range for each object property is defined. For instance, the object property algorithmType has a domain class Algorithm and has a range class AlgorithmicType.

**5.3 Data Property**
A data property is a property that links an entity with its property value. AMV consists of 46 data properties. Each data property is expressed with their respective domain classes and range data types. Of the total 46 data properties, 27 have Algorithm class as the domain. The rest of the data properties are of the other associated resources like Agent, Implementation etc. Some of the data properties are input, output, timeComplexity, edgeCase, averageCase, spaceComplexity, CPUTimeLimit, constraint, etc.

Figure 2 provides an overview of AMV ontology. It shows the classes, data properties, object properties and their cardinality restrictions. It can also be seen from the figure that AMV (re)uses the terms from the existing vocabularies, such as FOAF (http://xmlns.com/foaf/0.1/), Dublin Core Terms (DCT) (http://purl.org/dc/terms/), Schema.org (https://schema.org/), SKOS (http://www.w3.org/2004/02/skos/core#), DCAT (http://www.w3.org/ns/dcat#), and DOAP (http://usefulinc.com/ns/doap). For instance, AMV uses the terms from DCT are, Rights statement, License document, creator, title, created, description, identifier, etc. The terms from FOAF are Person, Organization, depiction, page, name, etc. The (re)use of terms from the other existing vocabularies is perfectly aligned with our stated principle in methodology section 3. This approach also makes AMV a linked data vocabulary.

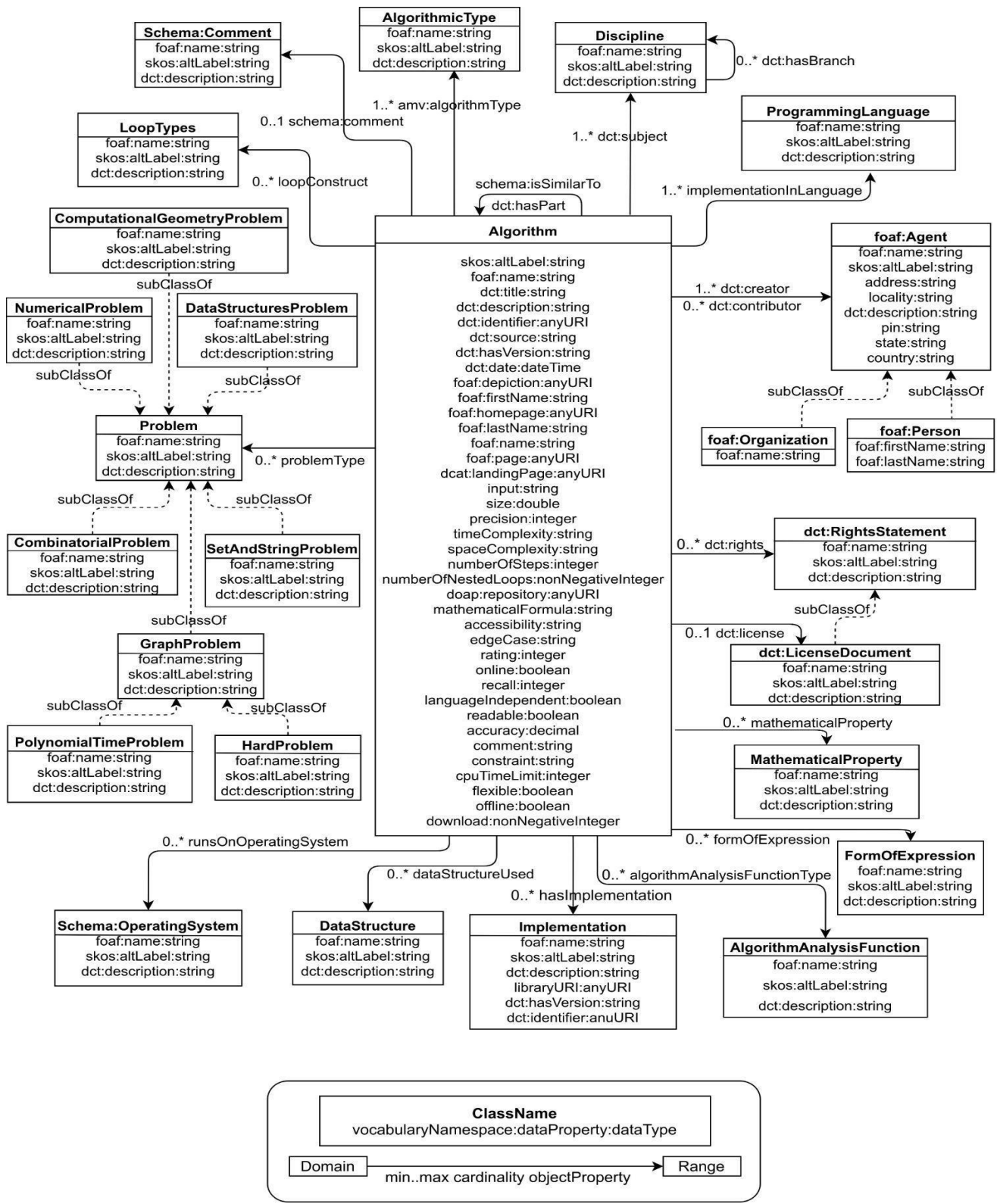

**Fig. 2.** AMV Overview. The terms with prefixes in the figure indicates the vocabularies that have been reused in constructing AMV. For example, "foaf" for http://xmlns.com/foaf/0.1/, "dct" for http://purl.org/dc/terms/, and "schema" for https://schema.org/. The terms without any prefix indicates that they are defined in the namespace of AMV i.e. https://www.isibang.ac.in/ns/amv#

## 6 Evaluation

The evaluation of AMV has been conducted in two different ways. The first one through a user survey and the second one by executing a set of SPARQL queries.

**6.1 User survey**

The survey led to comprehend the clients' algorithm search behaviour, what sort of data they normally look for, how and where they search, The review had 14 inquiries altogether, albeit in the study the inquiries are not classified. However, consistently the inquiries of the reviews can be assembled into three-wide classifications. As beneath. The review was coursed among the Professors in Computer Science, Ph.D. understudies, and programming experts.

Class 1: In this classification, we posed some fundamental inquiries from the clients planning to know their looking through conduct and the different areas where they lead their hunt be it the web or any repository. From these essential inquiries, we will have a thought regarding how a client lead's his/her inquiry and where they direct it.

Class 2: This classification addresses where we referenced a portion of the fundamental properties like the number of loops, serial complexity, time complexity, speed, etc and so on from our work, and found out if they think about these properties while looking for a algorithm. In the inquiries, we just demonstrated a portion of the essential properties displayed by the algorithms and not the intricate ones to keep away from disarray among the clients while noting the review. From these inquiries, we will actually want to break down the pertinence of our model with regards to ongoing use.

Class 3: In this classification, we had open-finished inquiries like what keywords the specialists of the area use while looking for an Algorithm, what filtering criteria they like in an archive, what administrations should an Algorithm repository give, what implementationsof the algorithms they search for. From these inquiries, we needed to remove a portion of the properties from the professionals, and check if we have them in our model.

**6.1.1 Survey Analysis**
Beneath we present the examination of the survey. This assists us with surveying our model and make it more concrete for ongoing use. The review was replied by a sum of 10 individuals a large portion of them working in the field for in atleast 5 years.

In the primary classification of inquiries, the client generally referenced that they search algorithms on the Internet, and a not very many of them allude to any repository for looking for a algorithm. 60% of the members reacted that they search on the web for Algorithm, 10% looked through the books and rest 30% referenced sites, similar to Stack Overflow and GitHub store. Figure 3 shows the reactions.

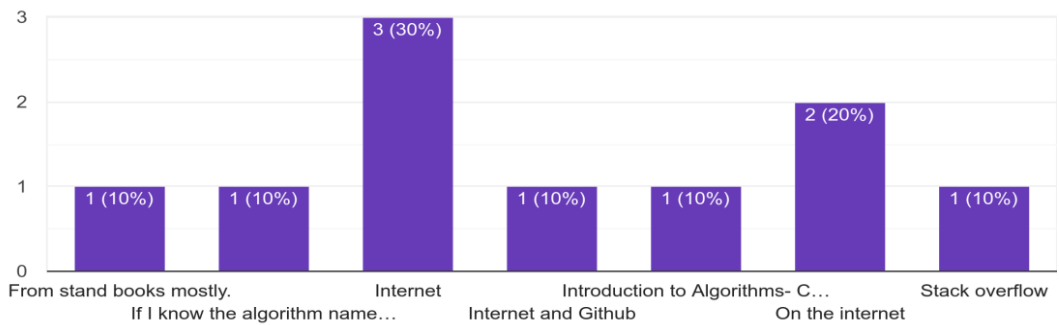

**Fig. 3.** Shows the response for the query "where do you prefer searching an algorithm? On the Internet, or in any specific repository?"

In the second category as we mentioned some properties from our model like the number of loops, hardware, time complexity, hardware requirements, applications of an algorithm. 100% of the users mentioned that they consider time and space complexity as one of the major properties of the algorithm. When asked about the number of iterations and nested iterations, 90% of the responses were in the favour that they do consider that in an algorithm.

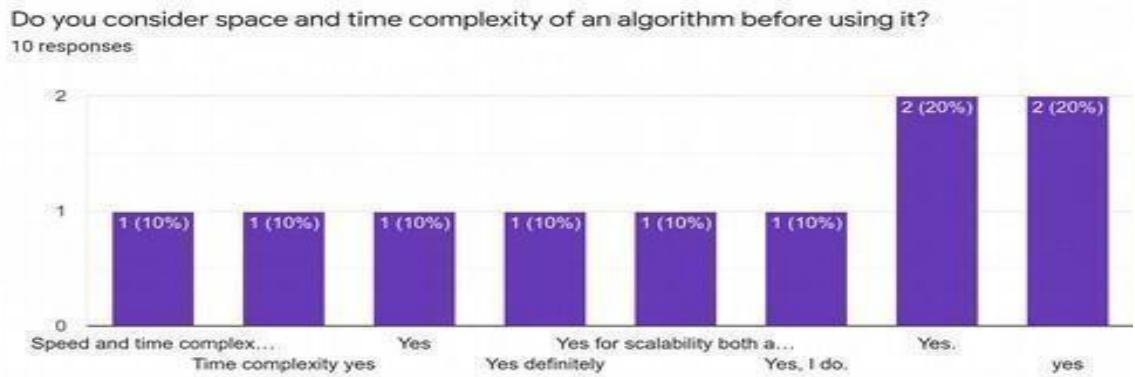

**Fig. 4.** Shows the response for the query "do you consider space and time complexity of an algorithm before using it?"

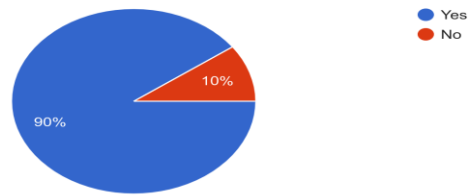

**Fig. 5.** Shows the response for the query "do you consider the number of loops and number of nested loops as one of the major factors for searching an algorithm?"

In the third class of inquiries, we posed open-finished inquiries. So we could extract some properties from the practitioners and match them with our model, the responses included that a user should be able to search in a repository with a purpose related keywords, like pattern matching, data structures used, sample input, time and space complexity, edge cases, implementation language, description of parameters. According to the users, the above mentioned properties should be provided in the description of the algorithm. All these responses have not only provided input in making our metadata elements concrete but also provided power to it for real-time usage.

The above-discussed survey and its analysis was one way of standardizing our AMV and making it concrete in real-time. Next, we present how our model will work in real-time and how it will answer the various queries of a practitioner.

**6.2 SPARQL Queries**

To assess AMV, we execute a bunch of SPARQL questions. The SPARQL inquiries are created by transforming the competency questions enrolled in procedure segment 3. The competency questions were created to show the reason for AMV. Subsequently, the fruitful recovery of wanted outcomes will demonstrate the adequacy of AMV and its semantic model. It is worth to see here that prior to running the questions, we populated the AMV information base with eleven algorithms and their property estimations. Figure 6 portrays one such algorithm and its property estimations delivered utilizing GraphDB [https://www.ontotext.com/].

Dutta, B. and Patel, J. (2021). AMV: Algorithm Metadata Vocabulary, preprint copy (prepared on April 29, 2021).

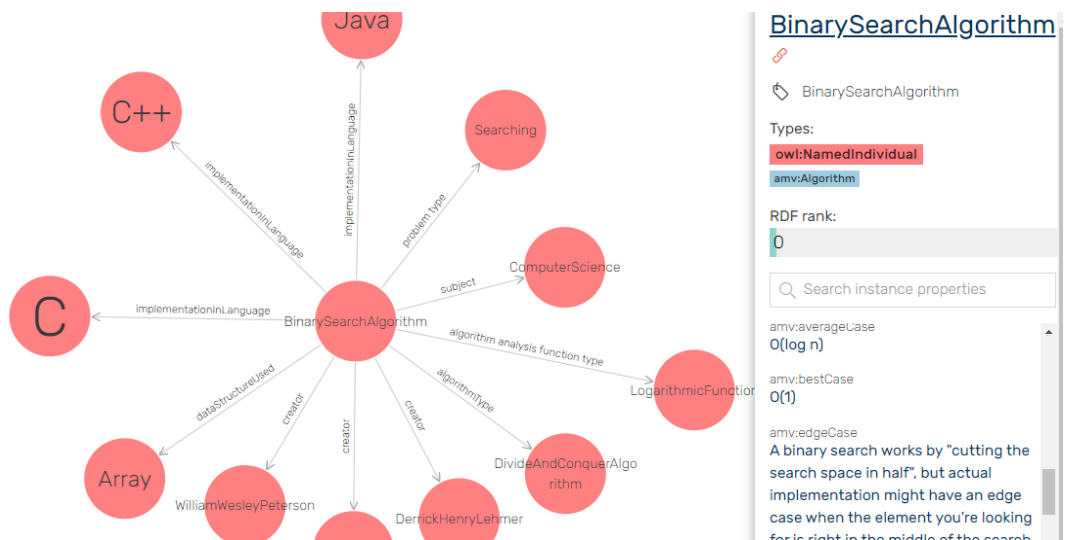

**Fig. 6**. Provides a visual of an algorithm from AMV knowledge base.

Figure 7(a) depicts the SPARQL query and the query result for a given competency question CQ2: Which language independent algorithm belongs to subject mathematics and solves combinatorial problems? Similarly, Figure 7(b) depicts the query and the result for a given competency question CQ3: Retrieve the list of algorithms that are implemented in python or java and have less than 7 steps? The successful execution of the queries shows the efficiency of the designed AMV vocabulary.

Fig. 7: Figure 7(a) & (b) depicts then SPARQL queries and results for the CQ2 and CQ3, respectively.

## 7 Related Work

As expressed, there exist no metadata metadata vocabulary for portraying the algorithm. The current work is the initial move towards this. In any case, as we probably are aware, there are many metadata vocabularies, beginning from generic metadata vocabularies (e.g., DC, DCAT, schema.org, MAchine Readable Catalog (MARC)) to the more article explicit vocabularies (e.g., Metadata for Ontology Description and publication(MOD), Friend Of A Friend(FOAF), Learning Object Metadata(LOM), ). Here, we momentarily talk about some of them.

Dublin Core schema is a metadata vocabulary that is used to describe physical and digital resources like books, CDs, web pages, images, etc. Dublin Core has two sets of metadata: the 15 core elements and in 2000 DCMI recommended some qualifiers to broaden the scope of the core elements called the qualified Dublin Core [1]. Schema.org aims to generate and preserve and promote schemas for structured data on the internet, on web pages, email, and more [3,4]. MARC standards are a set of digital formats for representation and communication of bibliographic and related information in machine-readable form [24]. DCAT is an RDF vocabulary to facilitate interoperability between datacatalogs published on the web [2]. Using DCAT to describe the datasets in catalogs, the publishers increase the findability and enable applications to absorb the metadata from several catalogs [17][2]. MOD is a metadata vocabulary for describing the ontologies, it uses equivalent terms from the many pre-existing metadata standards to ensure standardisation [5]. LOM is a data model, which describes the learning objects and the other digital resources which support learning. It is usually encoded in XML, the purpose of LOM is to reinforce the reusability of the learning objects and to provide better interoperability in terms of the online learning management system[23]. Schema.org intends to describe the structured data on the web for various entity types, such as Organization, Food, Person, Mind products, etc. DOAP (Description of project) is an RDF Schema and XML vocabulary for describing software projects. It was created to fetch semantic information related with open source software projects[21]. SKOS(Simple Knowledge Organisation System) is a model to demonstrate the basic structure and also allows to create relations between concepts in thesauri, subject heading lists etc[22]. FOAF(Friend of a Friend) A semantic web ontology for social connections between people, organizations and groups. It is an RDF- based vocabulary of classes of things and properties so that machines can process the information[20]. Some of these vocabularies, such as DC, FOAF, schema.org, SKOS, and DOAP are partially used in the current work as detailed in section 6.

## 8 Conclusion

Metadata is persuasive in extracting assets be it print materials or advanced resources like site pages, ontologies, algorithms, video, and then some. Aside from assuming an indispensable part in tracking down the different assets, it upholds the reusability of the asset. In this setting the current work is critical. As expressed in this paper, almost no work has occurred in overseeing and saving the algorithms. We accept that after the current work, there will be increasingly more work in the space beginning from metadata metadata vocabulary improvement to the committed algorithm store advancement. In our future work, we might want to expand and improve AMV by dissecting and adjusting it to the next important existing vocabularies, as PROV Ontology (http://www.w3.org/ns/prov) and Vocabulary Of A Friend (VOAF http://purl.org/vocommons/voaf), and so forth, likewise by offering accentuation to the different parts of the algorithm, for example, rendition control, execution, and so on Another significant exploration that we might want to do later on is the programmed extraction of algorithmic data from logical articles and for this reason, AMV is to be applied.